# t-STEP: An interpretable model for Total Electron Content predictions and irregularities estimations

**Stephen Tete[1*], Carl Shneider[1], Maxime Cordy[1], Claudio Cesaroni[2], Andreas Hein[1], Vasily Petrov[3]**

1 Interdisciplinary Center for Security, Reliability and Trust (SnT), University of Luxembourg, Luxembourg City L-1855, Luxembourg

2 Istituto Nazionale di Geofisica e Vulcanologia, Via di Vigna Murata 605, Rome 00143, Italy

3 Mission Space | Space Weather, Luxembourg City L-1191, Luxembourg

**Corresponding author:** Stephen Tete

Email: stephen.tete@uni.lu

Orcid ID: https://orcid.org/0009-0006-1881-7360

**Co-Author's Orchid**

Carl Shneider: https://orcid.org/0000-0002-3689-6959

Maxime Cordy: https://orcid.org/0000-0001-8312-1358

Claudio Cesaroni: https://orcid.org/0000-0003-2268-4389

Andreas Hein: https://orcid.org/0000-0003-1763-6892

## Abstract

Earth system infrastructures that utilize satellite-based informatics such as Global Positioning System (GPS) communications are regularly impacted by Ionospheric Total Electron Content (TEC) gradients. Unfortunately, modeling these TEC gradients under physical laws to guide mitigation strategies is challenging due to their dynamic and sudden occurrence likelihoods. For context, a multi-layer perceptron machine learning (ML) model could predict hourly TEC values to about 80% accuracy of a GPS receiver's estimates under solar wind and flux constraints. However, it is unclear whether these models could capture and preserve these TEC gradient irregularities within the predicted signals due to their output resolutions. To bridge this gap, we introduce, for the first time, an ML-based model trained to predict TEC at 30s resolutions. This cadence allows for the derivation of TEC rate changes (ROT and ROTI) as diagnostic irregularity indicators. The current implementation employs GPS observations across solar cycle 24, over a station located at 5.49 °S, 47.49 °W. A multi-metric evaluation framework, including dynamic time warping, is used to assess model robustness while SHAP (Shapely Additive exPlanations) enhances interpretability of feature attributions. For the 30s predictions, a test accuracy of 91% (MAE=4.38 TECU) is observed over the high solar activity period (2015), demonstrating the model's applicability for accurate TEC estimations. We benchmark TEC predictions against the International Reference Ionosphere (IRI-2020) and showed that the hourly variant of our model outperforms IRI by about 35% in accuracy, 57% drop in absolute errors and 54% gain in prediction skill. More importantly, we captured significant irregularity (ROTI) dynamics and morphologies with the 30s TEC model under three geomagnetic storm intensity levels in contrast to an attention based Long Short-Term Memory model subjected to the same sets of experiment. This study demonstrates the possibility of achieving scalable TEC irregularities detections with a single TEC-based model without explicitly training redundant models for individual transients.

## Contributions

1. First development of a high-resolution ML-based TEC prediction model
2. First detection of TEC irregularities from modeled TEC signals
3. Identification of key TEC variability drivers with SHAP analysis
4. Reproducible informatics workflow for TEC prediction, benchmarking and irregularity estimations

## Introduction

L-band radio signals transmitted from Global Positioning Systems (GPS) are highly susceptible to positioning errors. These errors usually result from sharp electron density gradients as the signals traverse a pool of free electrons, otherwise known as total electron content (TEC), within Earth's ionosphere. Here, two common effects are experienced by the trans-ionospheric EM signals, namely the phase advance (where the electromagnetic (EM) carrier leads the code by $\Delta t$) and group delay (the phase shift derivative with respect to frequency) mostly resulting in loss-of-locks, communication glitches, and meter-level positioning inaccuracies. For the end-user, these adversaries are intolerant for reasons that transcend impact on precision-agriculture, navigation, timing, remote sensing, military intelligence, and the global economy. Therefore, quantifying and predicting TEC is an important endeavor that requires meticulous attention to guide and inform mitigation strategies and decision making.

TEC is defined as the total number of electrons ($e^-$) within a $1m^2$ ionospheric cross-section integrated along a signal's ray path between the satellite and the receiver, where 1 TECU = $10^{16}$ electrons/$m^2$. The quantity exhibits variabilities in diurnal, seasonal, geographic, solar cycle, and geomagnetic activity (Liu et al. 2009; Seemala et al. 2023), making their predictions dynamic and challenging. TEC is very sensitive to ionospheric transients, making it a suitable quantity for early detection and forecasting of space weather and geological events such as earthquakes and tsunamis (Feng et al. 2023; Fuso et al. 2024).

The local time variation of TEC has been shown to impact trans-ionospheric radio waves, causing amplitude and phase scintillations in both low and high latitudes (Aarons 1993; de Oliveira Moraes et al. 2017; Juan et al. 2018; Sousasantos et al. 2023). In the past, Pi et al. (1997) have quantified these sharp TEC gradients using the rate of TEC index (ROTI). Following this proposal, several ionospheric transients have been easily classified and studied (de Oliveira Moraes et al. 2017; Cherniak et al. 2018; Carrano et al. 2019; Mrak et al. 2023; Martire et al. 2023; Ren et al. 2024; Vital et al. 2025). A spike in the rate of TEC (ROT) or ROTI could indicate an ionospheric phenomenon depending on its morphology, time of occurrence, and behavior. For example, a sharp transient on the scale of a few minutes could be related to a medium-scale traveling ionospheric disturbance (MSTID) (Kotake et al. 2006; Sivakandan et al. 2021; Fu et al. 2025) and can facilitate the evidence of spread-F phenomena in the mid-latitude regions (Paul et al. 2022). Similarly, a spike in ROTI in the low-latitude regions during post-sunset to post-midnight hours could be attributed to a plasma bubble or Spread-F in the mid-latitudes (Oliveira et al. 2020; Imtiaz et al. 2021; Carmo et al. 2023; Paul et al. 2024).

In the last five to ten years, the adoption of machine learning has revolutionized the way ionospheric TEC is modeled. For example, Tang et al. (2022) proposed a 2-day global TEC forecasting approach based on the Meta's Prophet machine learning algorithm. Habarulema et al. (2007) employed three (3) GPS derived TEC locations and used a neural network (NN) to predict TEC over the South African Geospace, outperforming the International Reference Ionosphere Model (IRI). Similarly, Nigusie et al. (2024) employed three gradient boosting and stacking machine learning architectures to predict the hourly values of the Vertical Total Electron Content (vTEC) where they achieved an accuracy of about 96% on the test distribution. Adolfs et al. (2022) developed a storm time NN model for predicting TEC over latitude 32.5°N - 70°N and longitude 30°W - 50°E, achieving a 35.6% improvement over the German Aerospace Agency's Neustrelitz TEC model (NTCM) during storm times and 17% increase during quiet time.

In Silva et al. (2023), a Multilayer Perceptron (MLP) regional TEC model was developed and validated across the Brazilian region, outperforming the NeQuick-G and Global Ionospheric Maps (GIM) by some 27% and 33% respectively in terms of error reduction. In the space of irregularities and ionospheric delay predictions, several works have applied data mining and machine learning techniques to reconstruct, as well as detect the likelihood of the occurrence of ionospheric irregularities given the dynamics of the geophysical space weather and ionospheric features (Rezende et al. 2010; De Lima et al. 2015; Sivavaraprasad et al. 2020; Nugent et al. 2021; Zhao et al. 2021). The application of machine learning for ionospheric analysis, nowcasting and forecasting has seen tremendous impacts especially when dealing with the enormous amount of data from the diverse state-of-the-art instruments.

However, a crucial gap remains regarding the ability of ML-based TEC models to retain irregularity morphologies within predicted TEC signals, without explicitly training specialized models on each of these transients. Both previous and current state of-the-art ML-based TEC prediction models have low prediction cadences, hence, beyond quantifying TEC modulations, it is practically impossible to estimate the state of TEC transients within them. In this work, we attempt for the first time to predict TEC at a 30s resolution. This cadence is synonymous with the sampling rate of geodetic receivers, hence making it relevant for irregularities estimations using established protocols. This time, we predict the TEC and compute the rate of change of TEC (ROTI) as a diagnostic indicator of irregularities. This approach also adds significance to the predicted TEC signals beyond the art of quantifying TEC modulations under different space weather activities.

## Data and Methods

### TEC, ROT and ROTI computations

In this work, our aim is to provide TEC estimations that are statistically significant for ionospheric irregularities detections. To aid the process, we use a ground-based GPS receiver located in Brazil $(5.49°S, 47.49°W)$. The station receives broadcast dual-frequency GPS signals at $f_1 = 1575.42\ MHz$ and $f_2 = 1227.60\ MHz$, allowing for the estimation of ionospheric delays, $(\tau)$.

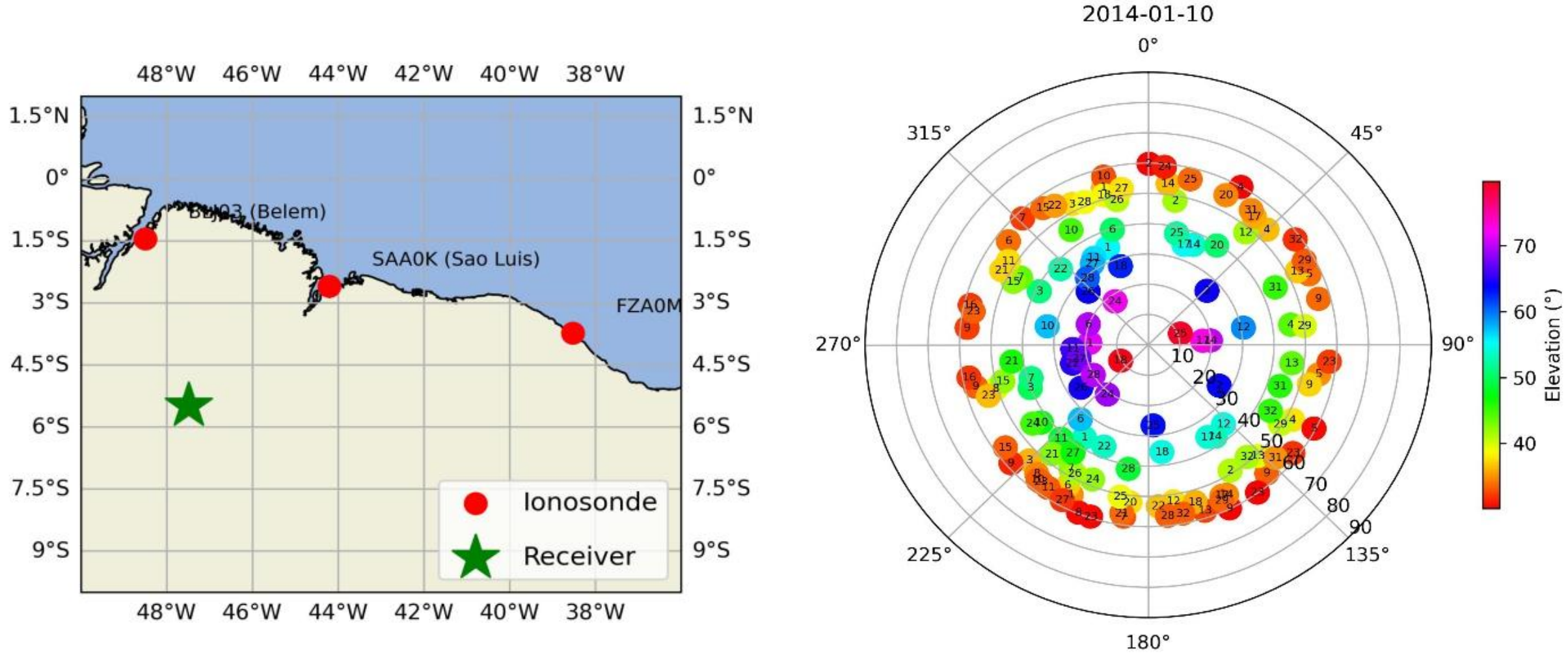

**Fig. 1** Geolocations of GPS receiver (green star) and ionosondes stations (red dots) used in this study are shown in the right panel, whereas a sky plot showing visible satellites in the field of the GPS receiver at their relative elevations on 10 January 2014 is shown in the left panel

$$\tau_i(f_i) = \frac{40.3}{c{f_i}^2} \cdot sTEC \tag{1}$$

We then compute the sTEC using Eq. 2.

$$sTEC = \frac{1}{40.3} \frac{f_1^2 f_2^2}{f_1^2 - f_2^2} (L_1\lambda_1 - L_2\lambda_2 + \varphi_n + \tau_n) \tag{2}$$

where c is the speed of light, $f$ is the signal frequency and $sTEC$ is TEC along the slant path of the signal or the slant TEC. $L_1, L_2$ are the carrier phase measurements at $f_1$and $f_2$. $\lambda_1$, $\lambda_2$ are the signal wavelengths whereas $\varphi$ and $\tau$ are the phase ambiguity and phase delays.

Assuming a thin-shell ionospheric layer at a height ($H_{ion}$) of 350 km, and a constant Earth radius $R_e = 3671\ km$, the obliquity function $M(\theta_{elv})$ was used to convert $sTEC$ to $vTEC$ (Vertical Total Electron Content) after which we apply an elevation ($\theta_{elv}$) mask of 30° to mitigate multipath, backscattering and tropospheric effects.

$$vTEC\ = cos\left(arcsin\left(\frac{R_e}{R_e+H_{ion}}cos\theta_{elv}\right)\right) \cdot sTEC \tag{3}$$

The rate of TEC change (ROT) was calculated using Eq. 4. We identified an ionospheric irregularity following the formulation of (Pi et al. 1997).

$$ROT_k{}^j = \frac{TEC_k^j - TEC_{k-1}^j}{t_k - t_{k-1}} \tag{4}$$

We adopted a sliding window of 5 minutes over ROT and computed the deviations for the values in the sequences to obtain ROTI. Throughout the proceeding experiments, this would be the formal derivation of ROTI after validations with existing literature and databases alike.

$$ROTI\ =\ \sqrt{\langle ROT^2\rangle - \langle ROT\rangle^2} \tag{5}$$

where $j$ is the visible satellite within the receiver's field of view, $k$ is the satellite epoch, TEC is the vertical total electron content and $\langle\rangle$ is the averaging operator.

## Data

Following the computation processes above, we obtained vTEC at each Ionospheric Pierce Point (IPP) for all instances of $j$ from 2009 to 2015 (7 years). In total, approximately $40 \times 10^6$ non-corrected TEC observations were obtained. After this, we used the closest ionosonde radar chain to the receiver (Fig. 1) to track the critical frequency (foF2) and virtual heights (hmF2). To ensure we had consistency with the GPS TEC observations, we mitigated several anomalies with the ionosonde datasets. First, the operational period of these ionosondes spanned 2012 to 2025, which was only a subset of the GPS receivers'. Secondly, the sampling rates are 10 minutes, and finally, there exist instances of complete signal absorption, making it impossible to measure the ionospheric properties. We will describe how these adversaries are mitigated in the modeling routine section below. We use the pyspedas framework Grimes et al. (2022) to obtain space weather parameters including Kp, SymH and Flow speed from the OMNI module to aid the modeling process. In its entirety, the potential feature space had about 36 variables (Fig. S1**)**. These would be trimmed to reduce model complexity and feature dimensionality.

**Modeling routine**

In this section, we describe the methods used to ensure temporal consistency with TEC and mitigate various instrument discrepancies before training. The routine begins by mitigating the major ionosonde data anomalies mentioned in the previous section. The 10 minute resolutions were resampled on the time index to match the vTEC measures. This introduced additional gaps on top of existing ones. A linear interpolation could solve this at the risk of losing intricate variability. This was not ideal. Hence, we resorted to an eXtreme Gradient Boosting (XGBoost) inference. At the time of data extraction, the Embrace and GIRO DIDBase databases had more than 13 years of ionosonde datasets running. This archive was enough to aid the development of XGBoost inference models for foF2 and hmF2 to cover the missing data.

Given that the sun directly controls ionization at atmospheric altitudes, the solar zenith angle (SZA) and solar radio flux at the 10.7 cm wavelength could be used to reconstruct foF2 and hmF2. For instance, at lower zenith angles the sun is nearly overhead, which implies direct heating, increased ionization and increased foF2 and vise-versa. The XGBoost sub-model uses year, day of year (Doy), time, F10.7 cm, Kp, SZA and sunspot number (SSN) to reconstruct the foF2 whereas for hmF2, we engineer a fractional feature (thus, ratio between SZA and F10.7) to aid the inference. This fractional feature, upon investigation, proved to be very important for hmF2 estimation rather than foF2. The foF2 and hmF2 would be used as physics constraints in the TEC model to guide irregularities occurrences. The following algorithm summarizes the ionosonde data reconstruction solution.

**Algorithm**: Routine for ionosonde ensemble reconstructions

**Input**: $\boldsymbol{x} \in \mathbb{R}^{(n \times d)}$
**Target with gaps**: $y \in \mathbb{R}^{n}$
**Gap solution**: $\hat{y}$
Let the observed set be, $\mathcal{O} \subset \{1, 2, 3 \dots \text{n}\}$
Let spread-F and open gaps be, $\mathcal{M} = \{1, 2, 3 \dots n\} \setminus \mathcal{O}$
**Model**: $f_{\theta}: \mathbb{R}^{d} \rightarrow \mathbb{R}$
Mapping: $y_i \approx f_{\theta}(\boldsymbol{x}_i), \forall_i \in \mathcal{O}$
**while** $\mathcal{M} \neq \emptyset$:
Loss: $\min_{\theta} \sum_{i \in \mathcal{O}} \left(y_i - f_{\theta}(\boldsymbol{x}_i)\right)^2$
Data Fusion: $\hat{y}_i = f_{\theta}(\boldsymbol{x_i}), \forall_i \in \mathcal{M}$
Reconstruction:
$\tilde{y}_i = \begin{cases} y_i, & i \in \mathcal{O} \\ \hat{y}_i, & i \in \mathcal{M} \end{cases}$
**end**

Earlier in the data section, we mentioned the feature dimensionality at our disposal. We manage this complexity through correlation analysis, mutual information gain, and post-hoc evaluations and retained ensembles that proved significant for the TEC prediction task. The retained ensembles were also consistent with findings in the literature. Ultimately, the matrix contained temporal covariates (year, time and Doy), satellite IPP information (Lat, Lon, Azimuth and elevation), Ionosonde parameters (NmF2, hmF2 and foF2) and space weather conditions (Kp, SymH, and Vx). SymH and Kp are maintained due to their respective storm indications at local (equatorial) and global levels. We decomposed Doy and time into their Fourier transformations using Eqs. 6 and 7 to encode seasonality and temporal cyclicality in the TEC model. In the end, we performed an ablation study to understand the impact of marginalizing the covariates on the models ability to predict TEC and capture irregularities within the predicted signal.

$$sdoy = sin\left(\frac{2\times\pi\times Doy}{365.25}\right), \quad cdoy = cos\left(\frac{2\times\pi\times Doy}{365.25}\right) \tag{6}$$

$$stime = sin\left(\frac{2\times\pi\times time}{24}\right), \quad ctime = cos\left(\frac{2\times\pi\times time}{24}\right) \tag{7}$$

**Model selection and architectures**

Several algorithms have been applied to predict ionospheric TEC and irregularities. A hand full of such algorithms that have seen tremendous successes includes, the gradient boosted decision trees, recurrent and neural networks. Recently, transformer architectures have also been explored. In this work, we have experimented with four networks namely: Long Short-Term Memory (LSTM), Light-Gradient Boosting Machine (LightGBM), Categorical Boosting and Random Forest and found that LSTM and LightGBM supported our objectives in terms of TEC prediction accuracy. LSTM is widely known, thus, going forward, we refer the reader to Yu et al. (2019) for more insight on the algorithm. This allows us to focus on the LightGBM as it has not been exploited in this area. Moreover, it is the main method adopted in this work due to its performance. However, we present a quick overview of the attention (ATT) technique used with the LSTM method (LSTM+attention). These two architectures provide a good basis for comparing the performance of a deep learning and a machine learning method.

**Light Gradient Boosting Machine (LightGBM) algorithm**

LGB. is a gradient boosting framework that is based on decision trees. It shines in terms of speed and efficiency and allows GPU acceleration. The algorithm uses an innovative Gradient-based One-Side Sampling (GOSS) algorithm to selectively retain gradients from large instances while randomly sampling those from smaller gradients. This is facilitated through leaf-wise growth and the selection of the leaf with highest loss reduction, helping to reduce computation cost and retaining accurate estimates of information gain for each node split using Eq. 8.

$$Gain = \frac{1}{2}\left(\frac{g_L^2}{h_L+\lambda} + \frac{g_R^2}{h_R+\lambda} - \frac{g^2}{h+\lambda}\right) \quad (8)$$

where $g, h$ are the sum of gradients and Hessians for all instances, $\lambda$ is the regularization parameter and $g_L, h_L, g_R, h_R$ are corresponding values for the left and right child nodes.

For a detailed breakdown of the algorithm, we refer the reader to (Ke et al. 2017). Under regression tasks, LGB tends to minimize an objective function over an ensemble of decision trees using Eq. 9. The loss is mostly a squared error, calculated using Eq. 10. In each sample, Eq. 11 and Eq. 12 are used to obtain the gradient and Hessian, which are then aggregated to estimate the optimal split points to update leaf weights. A prediction for sample $x$ is the sum of all tree outputs in Eq. 14.

$$obj = \sum_{i=1}^{n} l(y_i, \hat{y}_i) + \sum \Omega(f_t) \quad (9)$$

$$l(y_i, \hat{y}_i) = (y_i - \hat{y}_i)^2 \tag{10}$$

$$g_i = \frac{\partial l}{\partial \hat{y}_i} = -2(y_i - \hat{y}_i) \tag{11}$$

$$h_i = \frac{\partial^2 l}{\partial \hat{y}_i^2} \tag{12}$$

$$w_j = -\frac{\sum_{i \in I_j} g_i}{\sum_{i \in I_j} h_i + \lambda} \tag{13}$$

$$\hat{y}_i = \sum_{t=1}^{T} f_t(x_i) \tag{14}$$

where the regularization term, $\Omega(f_t) = \gamma T + \frac{\lambda}{2}\sum_{j=1}^{T} w_j^2$, $l(y_i, \hat{y}_i)$ is the loss function, $w_j$ and $T$ are the leaf weight $j$, and number of leaves respectively. $\hat{y}_i$ is the prediction for sample $x_i$ and $f_t$ is the output of tree $t$ for input $x_i$.

The algorithm tackles complex dimensionality by bundling mutually exclusive features and employs histogram-based learning to evaluate every possible split point, allowing for faster computation of split gains.

**Attention mechanism**

Attention mechanisms come in different forms including feature, temporal, multi-head, hierarchical attention and more. Amid these forms, the basic idea is to attend to strategic components that ensure generalizability. These solutions come in the form of assigning weights to the input elements to enable them to prioritize relevant updates within the modeling sequence. This is done by first computing similarity scores between each key $(K)$ and the query $(Q)$, then scaling by dividing the scores by the key dimension $(d_k)$. A softmax normalization is applied to obtain attention weights, which are then applied to the values for a weighted sum.

$$\text{Attention} = \begin{cases} scores = QK^T \\ \alpha = softmax\left(\frac{QK^T}{\sqrt{d_k}}\right) \\ Attention(Q, K, V) = \alpha V \end{cases} \tag{15}$$

In the current study, we applied feature and temporal attention to the sequences created for the LSTM. This is done to help the LSTM focus on contexts that are dynamically relevant for the current TEC predictions. While feature attention (instantiated with batch size, sequence length, feature dimension) applies weights to the features (example: SymH, V) at each timestep, the temporal attention produces a context vector (of batch size, lstm hidden dimension) to summarize the sequences with attention weights.

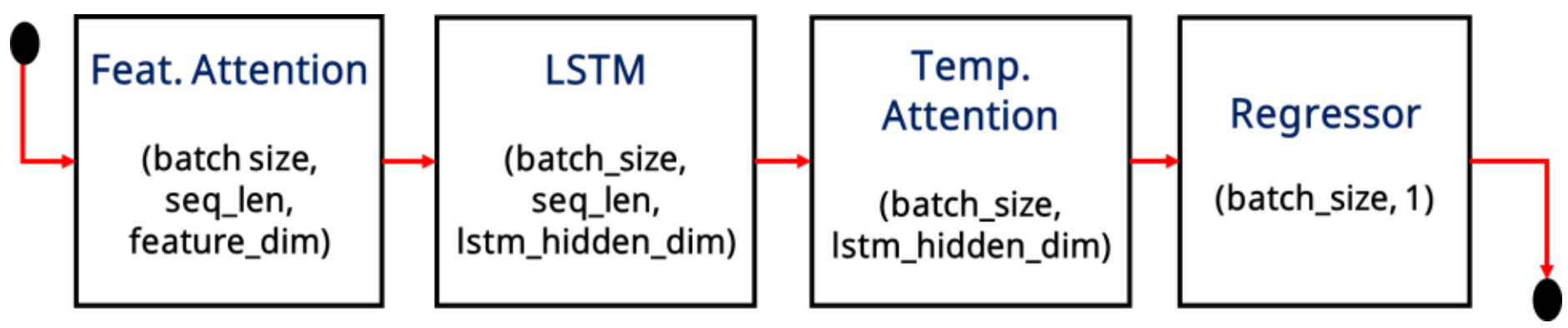

Following an exhaustive hyperparameter search using the same criteria for the LGB, we adopt (batch size, seq_len, lstm_hidden_dim, learning rate, early stopping patience, epoch) of (512, 720, 64, $5e^{-4}$, 5 and 100), respectively. Below is an implementation flow of the dynamic attention for the model used for the TEC prediction.

**Splitting strategy**

After preprocessing and correcting the data for any potential errors as well as applying an elevation mask of 30°, the final data was about $29 \times 10^6$. There are several ways of splitting datasets into training, validation and testing sets. In this work, we assume time dependence for training and testing and create a 6:1 year partition. Thus, 1 year (2015) was used for testing corresponding to about $3 \times 10^6$ whereas 6 years (2009 – 2014) was reserved for training and validation. To ensure that we validated across the solar cycle phases, we apply a 20% splitting and sampling from all phases and carefully inspecting for no leakages. Eventually, training samples were about $21 \times 10^6$ and validation set was about $5 \times 10^6$. We scale the data with the robust scaler to mitigate outliers.

**Hyperparameter tuning**

Due to the data size, it is expensive to run many ensembles. For this reason, we predefined sets of hyperparameters and identified them with run numbers. All runs were tested with subsets of the training, validation, and testing sets and monitored with the root mean squared error. We do not choose a set based on low TEC prediction errors but the one that improved irregularities retention after prediction. The main challenge was choosing the best learning rate and leaf numbers to optimize speed and accuracy. Finally, we settled on run 2 with 256 leaves and 0.1 learning rate. The run summary is shown in (Table S1) of the supplementary material**.**

### Model

To put our model in context, we refer to it as “t-STEP” (Temporal Sampling and TEc Prediction). Temporal sampling because, we systemically-engineered the routine to refactor its predictions on TEC, between 30s to 1h at the user’s request. For example, to compare with the state-of-the-art operational models, we refactor t-STEP to predict hourly estimates (t-STEP hourly) by varying the constraint requirements at observation time. Henceforth, any mention of “t-STEP” refers to our proposed TEC model. Figure 2 outlines the main routines discussed to this end whereas Fig. 3 describes the system’s architecture.

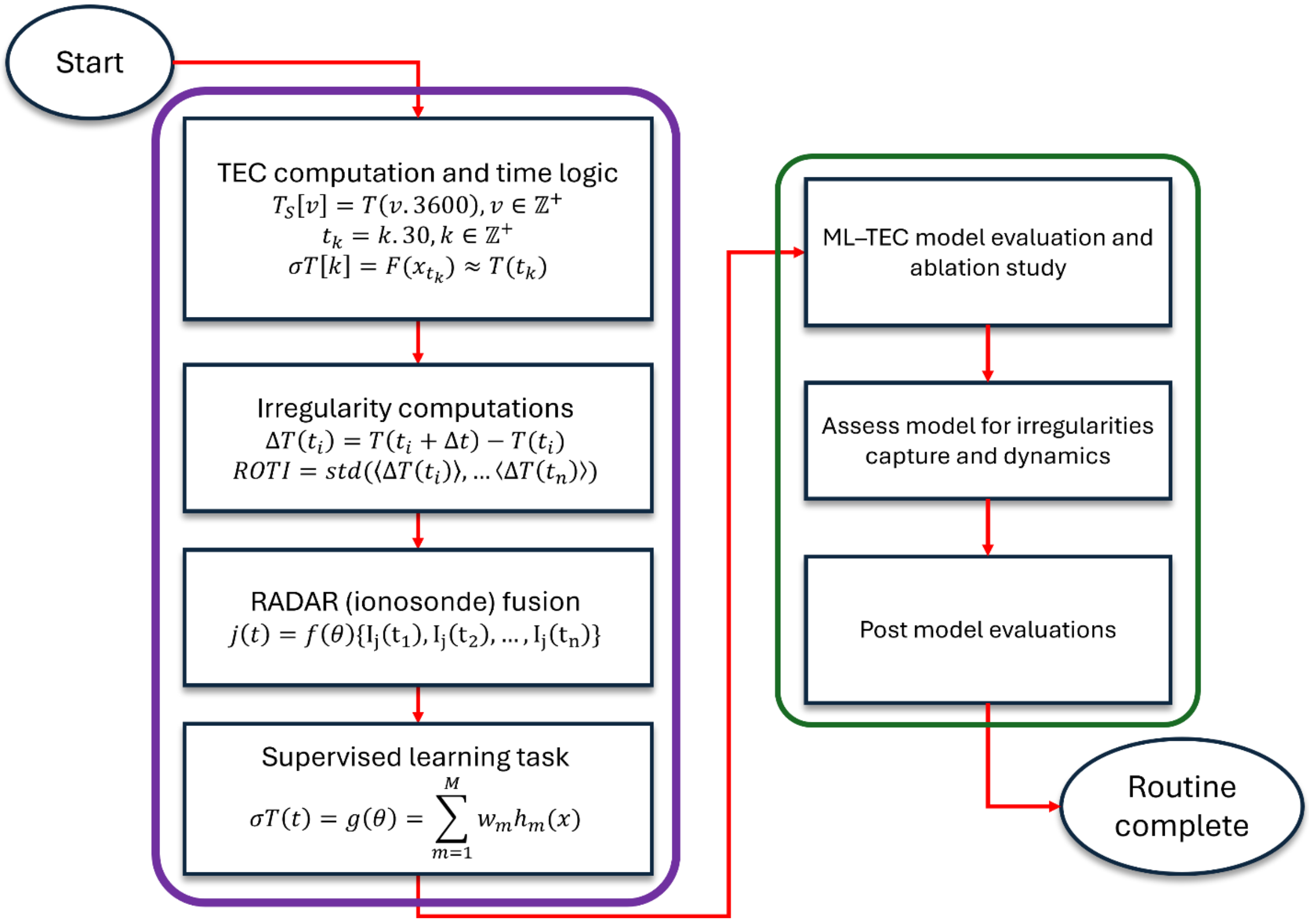

**Fig. 2** Flow diagram highlighting the steps and routines used to construct the t-STEP model. The violet rectangle summarizes the model's initial and development phases whereas the green rectangle represents the evaluation phases

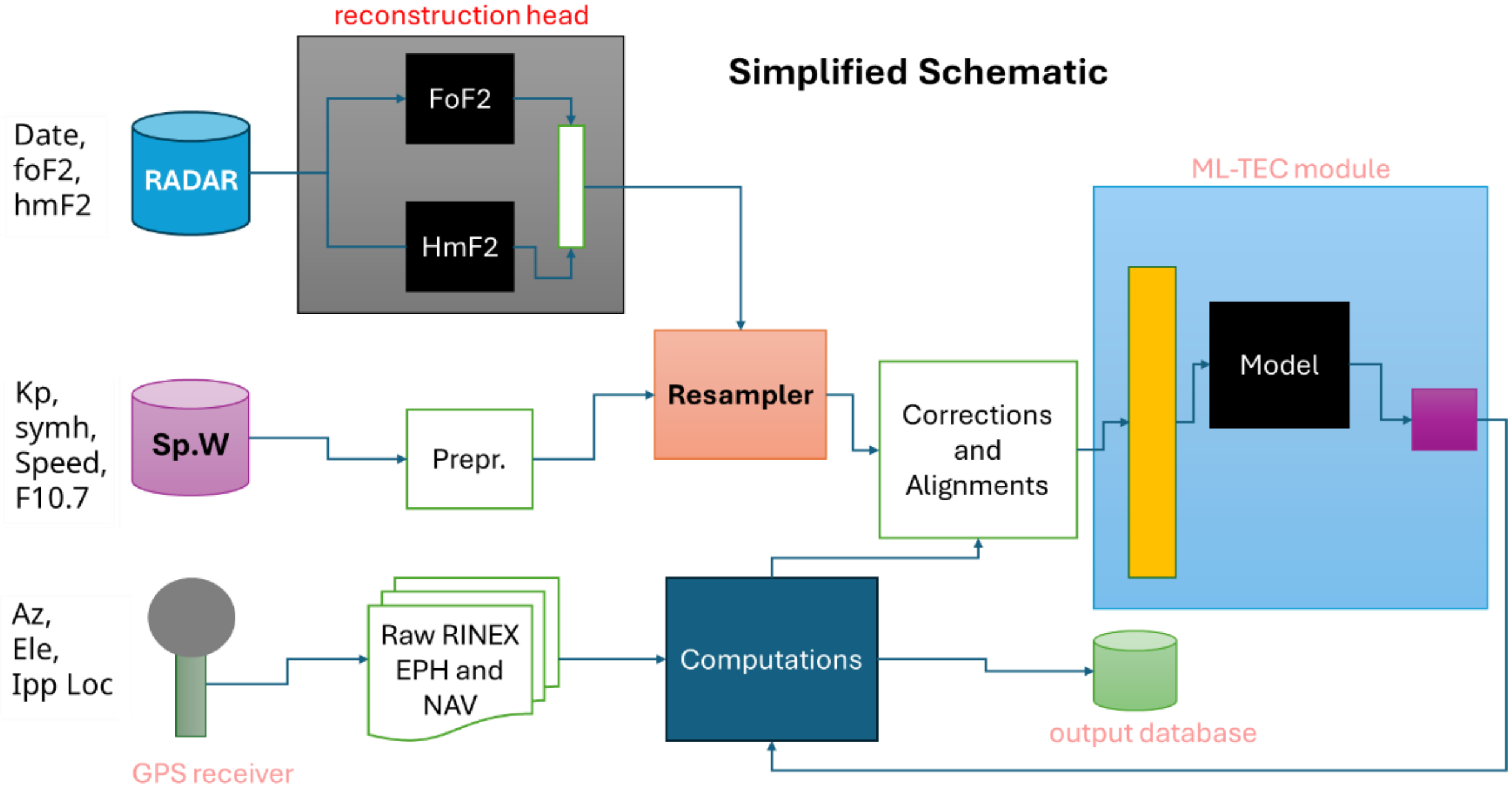

**Fig. 3** Simplified architectural representation of t-STEP's subsystems and routines.

## Experiments and model evaluations

The ionosphere responds to the dynamics of space weather variably. This means that the launch of ionospheric TEC irregularities is rather dynamic and does not always follow predictable patterns. For this reason, models that tend to predict these irregularities directly are easily overwhelmed by their nature, especially under different space weather conditions. As our approach is not direct, we gain impetus over this dynamic nature by predicting the TEC directly, after which we later infer the irregularity signatures from. Here, we set up one experiment under quiet, moderate, and severe geomagnetic activity conditions, respectively, and evaluate our model's performance towards TEC predictions and irregularity (ROTI) morphologies. For TEC and irregularity reconstructions, we benchmark our model against the LSTM and the International Reference Ionosphere (IRI-2020) model for TEC predictions.

We evaluated the models using the root mean squared error (RMSE), mean absolute error (MAE), correlation coefficient ($R^2$), concordance correlation coefficient ($\rho_{ccc}$) metrics. The metric ($\rho_{ccc}$) was introduced by Lin (1989). The metric measures how close the data points cluster, otherwise known as precision, and how they agree perfectly along the 45° line, thus giving a true representation of the alignment between the predictions and actual values in terms of location and scale. All other metrics have regular meanings.

$$metrics = \begin{cases} rmse = \sqrt{\frac{1}{N}\sum_{i=1}^{N}(y_i - \hat{y}_i)^2} \\ mae = \frac{1}{N}\sum_{i=1}^{N}|y_i - \hat{y}_i| \\ R^2 = 1 - \frac{\sum_{i=1}^{n}(y_i-\hat{y}_i)}{\sum_{i=1}^{n}(y_i-\bar{y})^2} \\ \rho_{ccc} = \frac{2\rho\sigma_y\sigma_{\hat{y}}}{\sigma_y^2+\sigma_{\hat{y}}^2+\left(\mu_y-\mu_{\hat{y}}\right)^2} \end{cases} \tag{16}$$

where $y_i$ and $\hat{y}_i$ are the observed and predicted vTEC data points, $\bar{y}$ is the mean of the observed values. $\rho$ and $\sigma$ are the Pearson correlation coefficient, and standard deviation whereas $\mu$, $\rho_{ccc}$ and $N$ are the mean and concordance correlation coefficient and number of data points in a given population.

**Dynamic time warping (DTW)**

DTW is a significant dynamical algorithm used to perform similarity checks between two time evolving signals, such as time series that may differ in phase, length, pulse locations, or speed, which is better than instantaneous distance metrics such as the Euclidean distance. DTW is different in such a way that it can warp the time axis to match similar patterns, shapes, and magnitudes (if they occur) at different rates.

Given two signals, $x_i, x_j$ where, $x_i \in x_{i1}, x_{i2}, \dots, x_{im}$, $x_j \in x_{j1}, x_{j2}, \dots, x_{jn}$. The pairwise distances are computed and the warping path, also given as the path through the cost matrix, $D \in \mathbb{R}^{(n+1)\times(m+1)}$ that minimizes the overall cumulative distances is estimated. After this, the DTW distance is the minimum cumulative cost required to traverse the optimal path and align the sequences of $x_i$ and $x_j$.

$$D_{i,j} = d(x_i, x_j) + min\begin{cases} D_{i-1,j-1} \\ D_{i-1,j} \\ D_{i,j-1} \end{cases} \quad (17)$$

Due to the adaptability of this algorithm, several proposals have been made to use DTW as a loss function to train machine learning models (Cai et al. 2019). This algorithm also finds practical applications in timeseries forecasting, where forecast horizons shift the series in time, relative to the ground truth.

We understand that the temporal positions of TEC transients within the machine learning model may be shifted either by early or later predictions. This is the case for all empirical models and may have an effect on the morphologies after reconstruction. In practice, evaluating such scenarios under regular linear metrics may not reflect full assessments of the model. Hence, we utilize the DTW algorithm to augment these linear evaluations.

**Comparison of t-STEP with the International Reference Ionosphere (IRI) model**

By 2014, the International Reference Ionosphere (IRI) has been accepted by the International Standards Organization as the standard for ionospheric parameterizations (Bilitza et al. 2014). It provides critical parameters such as Temperature, electron density, TEC, foF2, hmF2, and ion drifts at atmospheric altitudes (50 to ~2000 km) to support operational needs. For this reason, we chose to measure the improvement level by validating it against t-STEP. The output cadence of the IRI TEC is about 1 hour; hence, we use the hourly output of t-STEP to ensure consistency. Finally, the multi-metric evaluations Eqs. 16 and 18 are used to assess the experimental performance.

$$SS = 1 - \frac{E_{t-STEP}}{E_{IRI}} \quad (18)$$

where E is the error with respect to t-STEP and the IRI models, and SS is the skill score.

**Section summary**

In this section, we have shown how TEC and ROTI were computed to attain the 30s observation series. We have described the datasets and methods used to harmonize inconsistencies with them, including routines used to mitigate observations from the different experimental instruments. Feature curation and model selection criteria and schematics have been discussed. Finally, we present the experiments, evaluations, and benchmarking strategy against the international reference ionosphere, to quantify improvements in our model.

## Results and discussions

In this work, we performed several experiments and model tuning aimed at obtaining a model that predicts the ionospheric TEC at a higher temporal resolution, relevant for estimating ionospheric irregularities or transients. We must state that we employed a single station receiver approach at this stage. To the best of our knowledge, the model presented in this paper is the first of its kind under the single station approaches that have been trained with an enormous database due to the method adopted.

Seven different variants of the t-STEP model were tested, among which one was selected based on its ability to generalize and capture significant ionospheric transients translated in the ROTI under varying space weather conditions. Table 1 presents the metrics for the adopted version of t-STEP, whereas the other variants will be presented under the ablation study.

**Table 1** Performance quantification of t-STEP in different stages of model development

| Model | Modeling Stage | MAE (TECU) | RMSE (TECU) | $\rho_{ccc}$ | $R^2$ |
|---|---|---|---|---|---|
| | Training | 2.46 | 3.34 | 0.98 | 0.96 |
| **t-STEP (30s)** | Validation | 2.46 | 3.33 | 0.98 | 0.96 |
| | Testing | 4.38 | 5.75 | 0.95 | 0.91 |
| | Test on hourly cadence | 3.89 | 5.10 | 0.97 | 0.93 |
| **t-STEP hourly** | t-STEP | 3.92 | 5.25 | 0.97 | 0.93 |
| | IRI-2020 | 9.06 | 11.36 | 0.79 | 0.69 |
| | Skill Score | 0.57 | 0.54 | N/A | N/A |
| **LSTM + ATT** | Training | 2.94 | 3.91 | 0.97 | 0.94 |
| | Validation | 4.31 | 5.81 | 0.95 | 0.91 |
| | Testing | 4.72 | 6.21 | 0.95 | 0.89 |

Table 1 shows the metrics of t-STEP and its hourly variant, benchmarked against IRI-2020 and the LSTM deep learning model. Overall, we observed minimal prediction errors with t-STEP on the 30s TEC values. For instance, across the entire test population, an absolute error of 4.38 TECU was recorded, which averages to about 7.2% error reduction in t-STEP over the LSTM.

Similarly, the hourly variant outperformed the IRI-2020 model, gaining ~35% in $R^2$ accuracy, ~57% drop in MAE and a skill score of ~54%, demonstrating the operational capability of t-STEP. Relative to the training and validation accuracies, the test accuracy fairly drops. This drop is understandable especially when the test population is in the maximum (peak) phase of the solar cycle (2015). The ionospheric conductivity becomes strong with heightened fluxes and increased frequency of strong TEC gradients. Nevertheless, the error ranges are adequate and low.

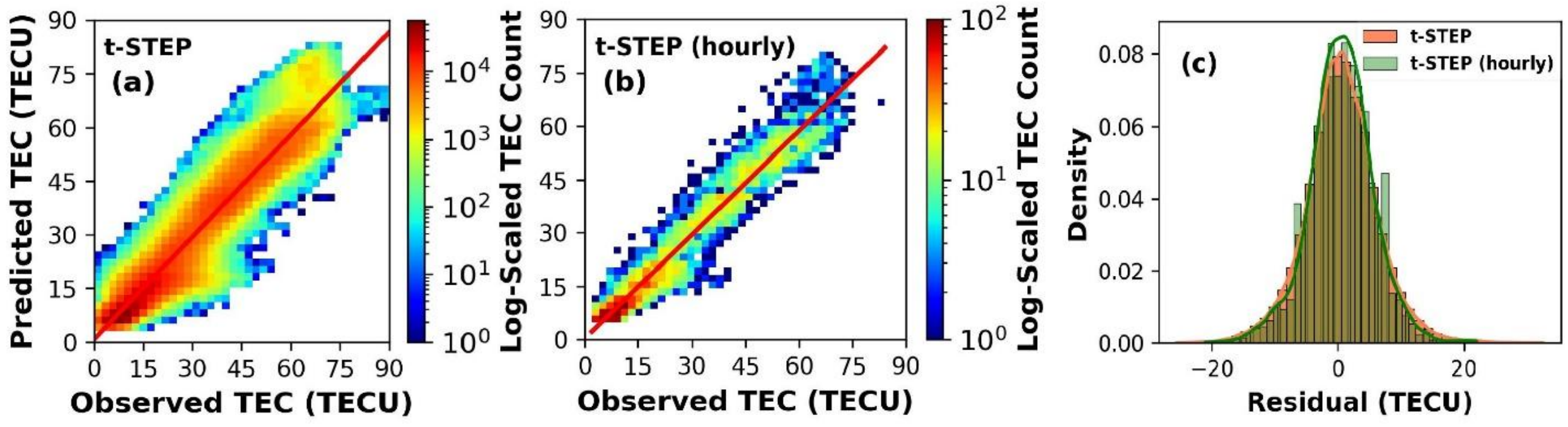

**Fig. 4** Regression between observed and predicted TEC on the test population for t-STEP (a) and the hourly variant (b). The distribution of residual errors for the two plots (a) and (b) are colored orange and green, respectively (c)

Figure 4 presents the regressions and residual distributions for the t-STEP and t-STEP hourly models. The general prediction ability of the model, as confirmed in Table 1, is great, especially in predicting most of the absolute TEC magnitudes. We observed that the residual distribution is nearly gaussian, with most of its values concentrated near zero. In Fig. 4a, we see that the dynamic ray paths of the satellites give the TEC broad distributions. Further investigations revealed that this stems from the geomagnetic strength disparities at the location of the GPS receiver. Ray tracks traversing regions of the geomagnetic equator experience dynamic elevated TEC magnitudes and strong temporal gradients compared to off magnetic equator regions due to the differences in electrodynamics. Without averaging these effects on the TEC as is done with hourly TEC models, t-STEP still predicted the TEC very well (4a) and even better when averaged (4b), asserting the model's robustness to external factors and capturing the underlying physics seamlessly.

## Seasonal variations

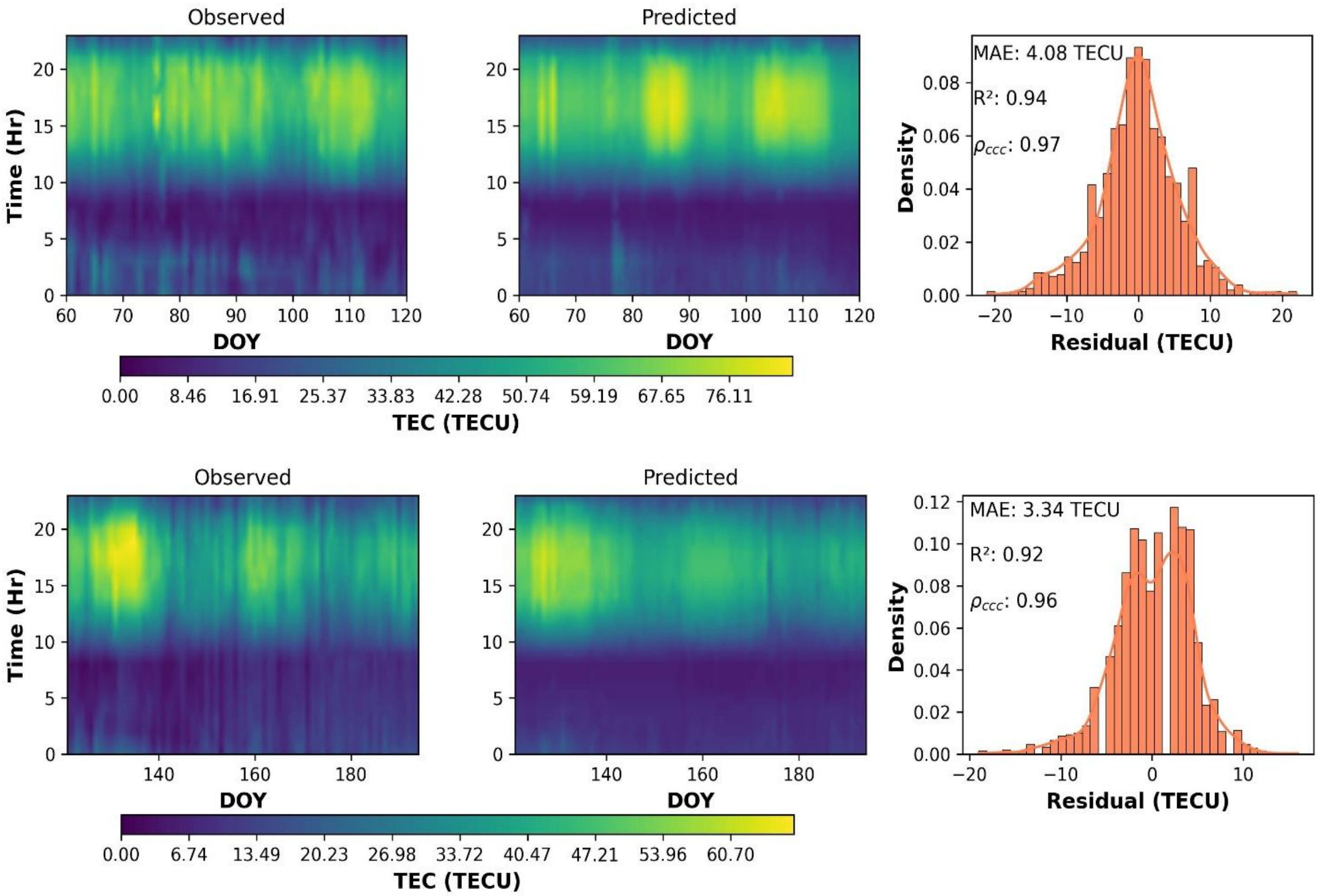

**Fig. 5** Seasonal plots showing t-STEP's variability in two seasons of Autumn (top panels) and winter (bottom panels) and their associated residual distributions (right panels)

In Fig. 5, the observed (left) and model reconstructions (middle) have been grouped into two seasons in the Brazilian sector. March to April (Autumn) in the first row and May to July (Autumn – winter). The model's reconstruction of the seasonal TEC morphologies is also phenomenal, showing consistent variabilities. Observed patterns included: higher TEC values in Autumn than in winter, weaker troughs between 4 am and 9 am, and higher crests as the sun rises. These are consistent with the findings of (Venkatesh et al. 2015; Dias et al. 2020). The residual distributions in the right panel of the figure confirmed the strong ionization background in March – April that contributed to the slight error increments (MAE = 4.08 TECU) than in winter (MAE

= 3.34 TECU). More importantly, t-STEP does not overestimate the TEC values nor significantly underestimate them in a way that may affect the morphology of any intrinsic transient.

## Evaluating t-STEP's robustness on space weather dynamics and irregularity detections

In this section, we investigated t-STEP's ability to capture TEC irregularities using the ROTI parameter. Since ionospheric irregularities are dynamic under quiet and active space weather conditions, we selected three different levels of geomagnetic activity (i.e. weak-to-negligible storm, moderate and strong storms) to ensure holistic evaluations. To guide this selection, we rely on the storm classification thresholds of (Gonzalez et al. 1994; Chakraborty and Morley 2020) and Space Weather live. We must emphasize that vTEC is the only predicted parameter. ROTI is computed using the formulations in Eqs. 4 and 5 as a diagnostic indicator of model robustness.

### Weak-to-negligible storm conditions

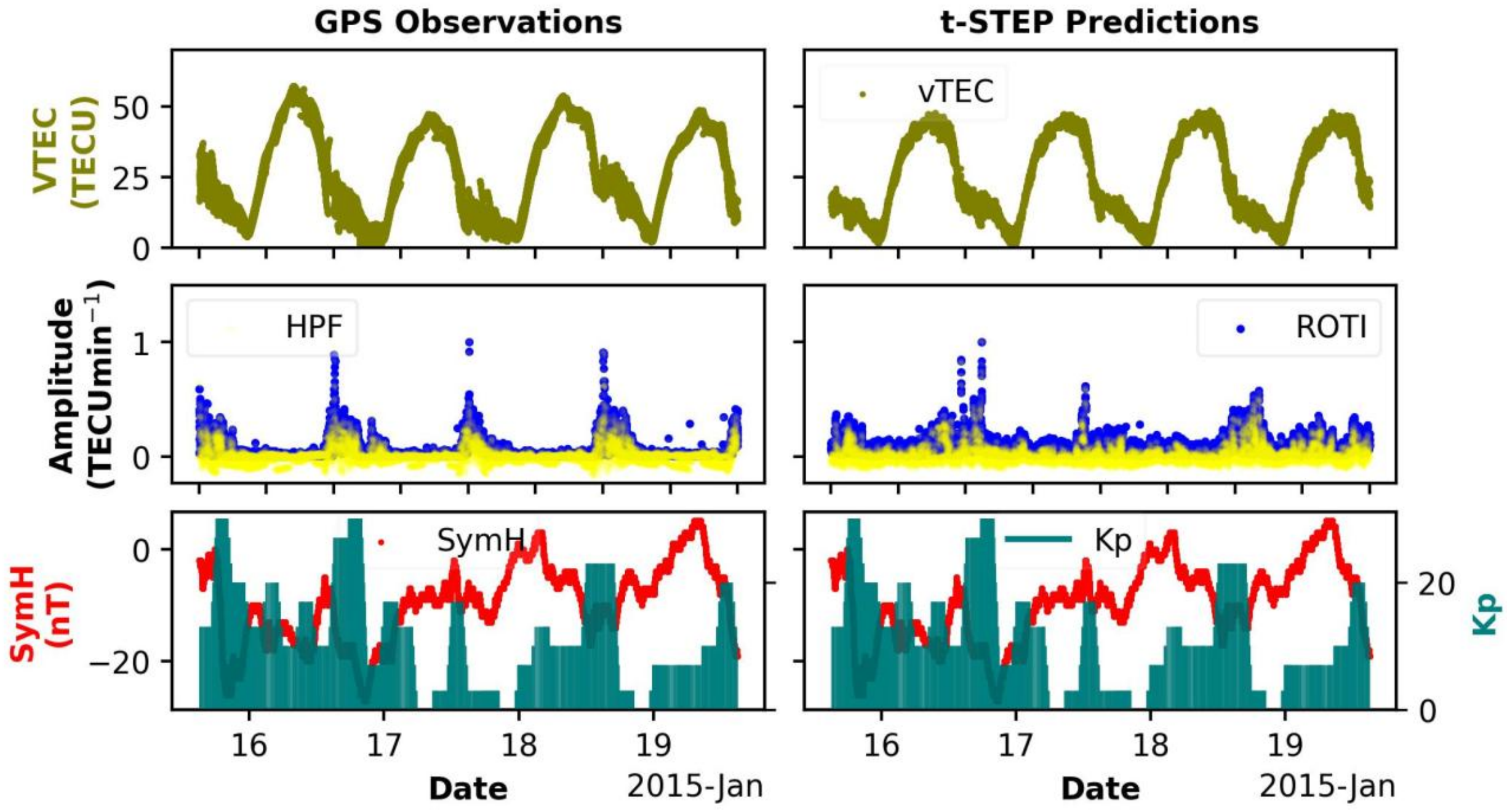

**Fig. 6** Comparison between GPS TEC and t-STEP TEC (top), ionospheric irregularity indicator (ROTI) (middle) and geomagnetic indices (Kp and SymH) during a period of weak-to-negligible storm conditions (bottom panel)

The left panels in the first row of Fig. 6 correspond to GPS observations, while those on the right represent t-STEP's predictions. The second row shows the diagnostic ROTI reconstructions. The two storm-time indices in the bottom panels describe quiet space weather conditions with SymH reaching -30 nT and a Kp of 3. In the first row, t-STEP's output is compared with the GPS observations, where we observed good TEC agreements throughout the series, signaling the model's adaptability under quiet time conditions. The model's TEC output tracked the overall temporal dynamics of the GPS observations and reproduced the diurnal TEC structure closely with $R^2, \rho_{ccc} = 0.90, 0.95$ (see residual distributions in Fig. 9a).

Interestingly, both GPS and t-STEP TEC preserved consistent ROTI morphologies and timeliness (middle panels). This observation explains the extent to which the model understood the physical conditions that favor the formation of ionospheric irregularities. The ROTI irregularities were mostly constrained at post-sunset and post-midnight hours and occurred on all 4 days of the observation period. Since our interest is primarily on the large irregularities, we passed the two signals through a Butterworth high-pass filter (HPF) to analyze the high frequency signals from both observations (yellow plots). This gave clear indications of periods within our model's product where improvements may be needed. A typical example is seen around noon on January 19, where negligible irregularities were seen on GPS, slightly contrasting the model's estimation. We must clarify that ROTI cannot be negative; hence, both GPS and t-STEP have unconstrained minimums of 0, except for the filtered signals. A detailed quantitative assessment of ROTI signals is provided in a later section.

**Moderate-to-weak Storm conditions**

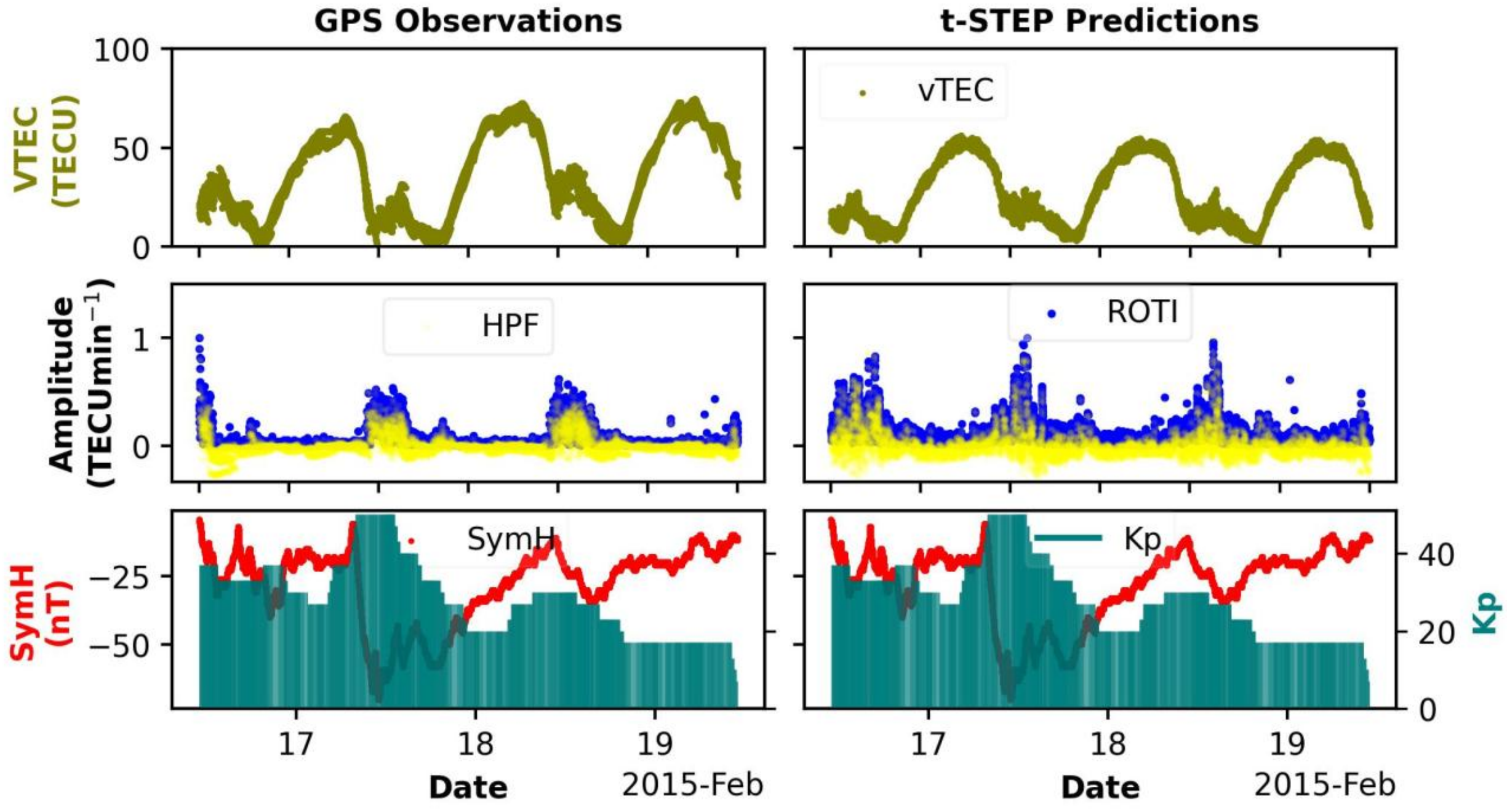

**Fig. 7** Same as Fig. 6, except for a period of moderate-to-weak storm activity, indicated by the Kp and SymH indices (bottom panel)

Figure 7 takes the same form as Fig. 6, except that the geomagnetic activity level is considered moderate following a weak storm that occurred before noon on 17 February. The second storm's sudden commencement (SSC) hit Earth before 20:00 hours on the same day, but the minimum excursion was seen on the 18th. Here, the minimum and maximum values of SymH and Kp were about -70 nT and 5. An increase in TEC was observed later in the day with significant suppression of conditions that triggered storm time irregularities until after the sun set. With this storm, it can be observed that t-STEP tends to under predict the main phase TEC modulations, which translated in $R^2$, $\rho_{ccc}$ of 0.73, 0.84 (Fig. 9b). This underprediction might have resulted from the multi-step-wise storm injection nature (see SymH in the bottom panel), which appeared to complicate the general storm-time electrodynamics learned during training.

Nonetheless, the TEC irregularity law that governs the model appeared unbroken. Adequate large scale morphologies and variabilities were maintained with fewer false signals than during the negligible storm conditions. These observations assert confidence in t-STEP's output, highlighting its robustness to physics and presenting a good balance between generalizability and accuracy.

**Strong-to-moderate storm conditions**

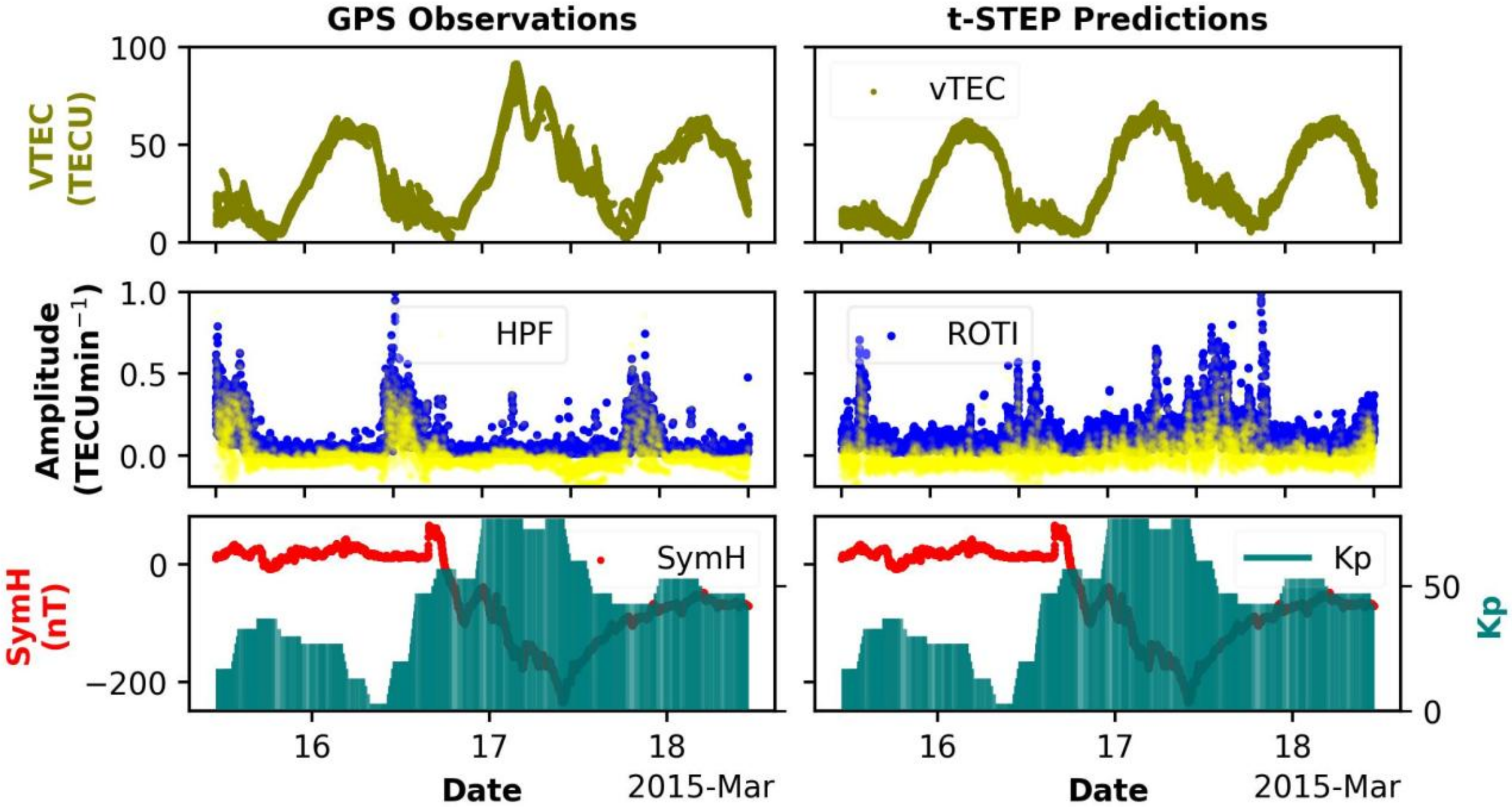

**Fig. 8** Same as Fig. 6, except for a period of strong storm activity (Patrick's day Storm), as indicated by the Kp and SymH indices (bottom panel)

One of the famous geomagnetic storms in the first half of the third millennium has been the 17 March 2015 storm. This storm, also known as the "St. Patrick's Day Storm", made headlines due to its major two-step morphology and strong impact on interplanetary space and at Earth. Several literatures have termed it a super storm (Wu et al. 2016; Suji and Prince 2018) due to the level of compression-driven disturbances at Earth. The storm disturbance index (SymH) hit below -200 nT and Kp recorded $8^{-}$. The impact of Patrick's day storm was felt on several infrastructures, including disrupting Railway (Thaduri 2020), Power grids (Carter et al. 2016), GPS and precise positioning (Paziewski et al. 2022) operations.

In Fig. 8, we observed several storm effects, including an increase in ionospheric TEC after the first storm impact, which signifies the injection of more electrons later into the ionosphere. This general trend was captured in the model, including the interesting daytime TEC bite-out phenomenon (except that the magnitude was small) at around 15 hours on the 17th. Here, we observed a prediction accuracy of $R^2, \rho_{ccc}$ $0.88, 0.94$ (Fig. 9c). An interesting contrast between the predictions in Figs. 7 and 8 is in the metric dynamics. One would expect strong accuracy drops under the strong space weather conditions compared to the moderate. However, we observed otherwise. For instance, given the same observation duration (3-days), the absolute error changed by ~77% with $R^2$ accuracy drop of ~17% (metrics inferred from Fig. 9) between the strong and moderate storm conditions. This happened because the storm morphology in Fig. 7 seemed more complex. There existed about six (6) more excursions (see SymH) of different storm magnitudes within the observation window than the two (2) observed in Fig. 8. This goes to confirm that duration-driven storms have more severe impacts than intensity-driven storms, both on models and on infrastructure.

The TEC irregularities in Fig. 8 have also been reconstructed in a way that gives the irregularities alerts in response to the storm-time gradients. The ionosphere over the GPS receiver responds to geomagnetic storms at some $\delta t$. The peculiarity of the Patrick's day storm (i.e., two storms with the second hitting before the first recovers), triggered a different irregularity occurrence dynamics in both GPS and t-STEP. As the first storm struck on the 17th, irregularities were suppressed by injections from the second strongest storm. The irregularities re-emerged during the day on the 18th, before noon. These observations are consistent with the physics of storm time irregularities, underscoring the sensitivity of both systems to storm timing and sequencing.

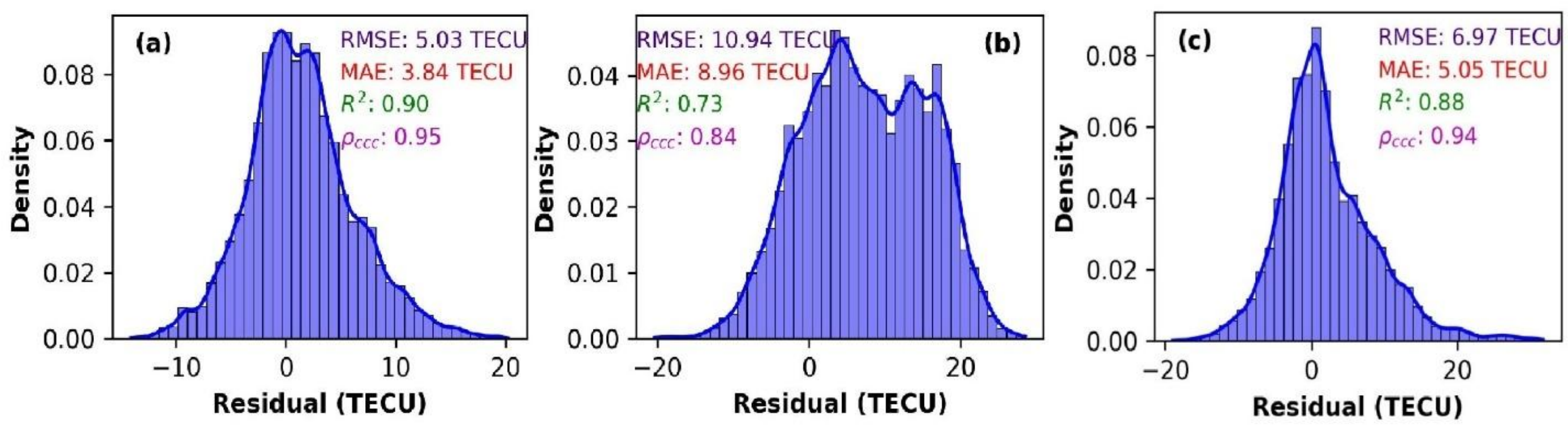

**Fig. 9** TEC residuals plotted as distributions for the three storm cases. Panels (a), (b) and (c) correspond to observations shown in Figs. 6, 7 and 8, respectively

## Assessing irregularity reconstructions with power spectral density (PSD) and dynamic time warping (DTW)

Although ROTI was only used to understand the retentive ability of the empirical models to embedded transients, it is important to understand them as they described significant ionospheric events that are usually studied independently. During the analysis, we discovered a limitation in the t-STEP's reconstructions. There are few instances in which irregularities appeared early or late in the TEC signals. Due to this, we performed a power spectral density (PSD) analysis to assess t-STEP's product to understand the likely causes (Fig. 10).

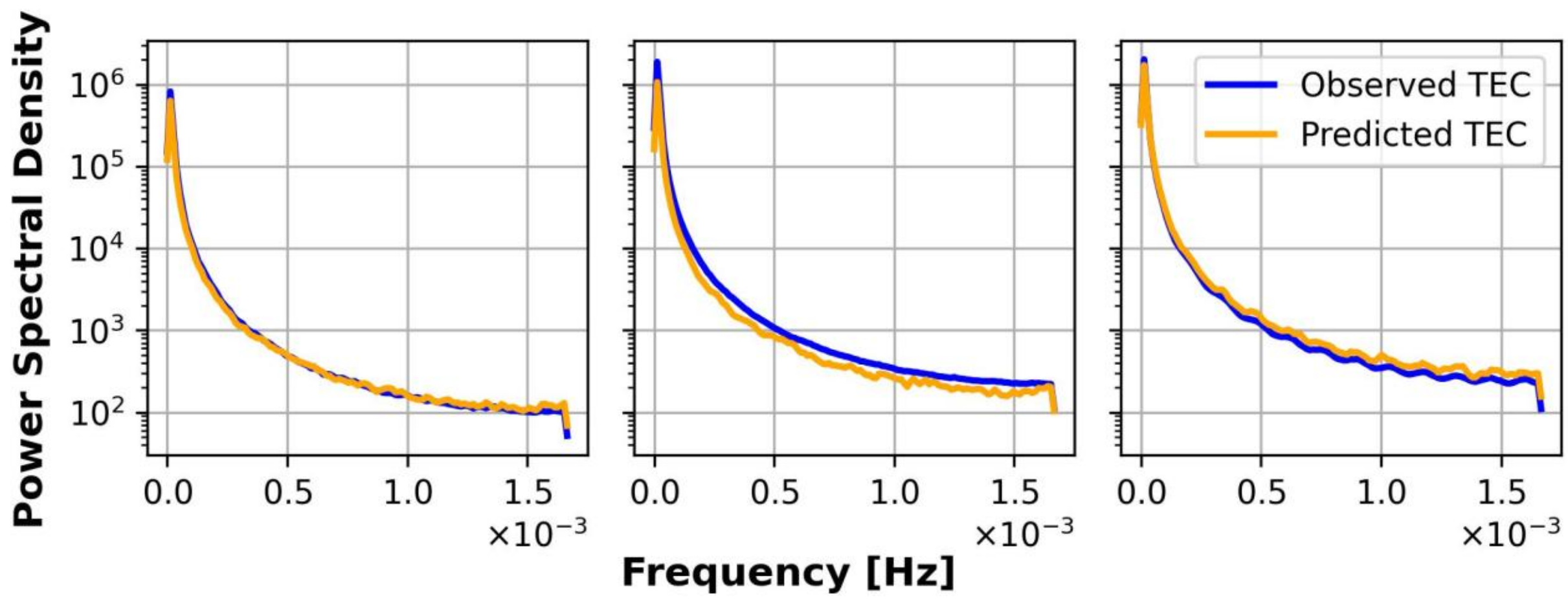

**Fig. 10** TEC power spectral density (PSD) plot for GPS (blue) and t-STEP (orange) under the three storm scenarios presented in Fig. 6 (left), Fig. 7 (middle), and Fig. 8 (right)

The PSD analysis in Fig. 10 confirmed the metrics in Fig. 9 and showed that the moderate storm (Fig. 7) induced the largest gradient divergence from the GPS (middle panel). The temporal discrepancies appeared to result from high frequency variations. This behavior highlights the impact of storm-time constraints on model operations. In all scenarios, variability and overall trend remained consistent with peculiar stable behaviors at low frequencies that highlight the model's robustness.

### Ablation study

In this section, we marginalize the contributions of the features and analyze their effects on the model's performance. The features were grouped and assigned a unified label based on the instrument of measurement. The groups had the following names and definitions: IPP (Lat, Lon), Geom (Kp, SymH, Flow speed), Iz (FoF2/NmF2, hmF2) and strategic combinations. This allowed the analysis to focus on both instrument level and feature behaviors.

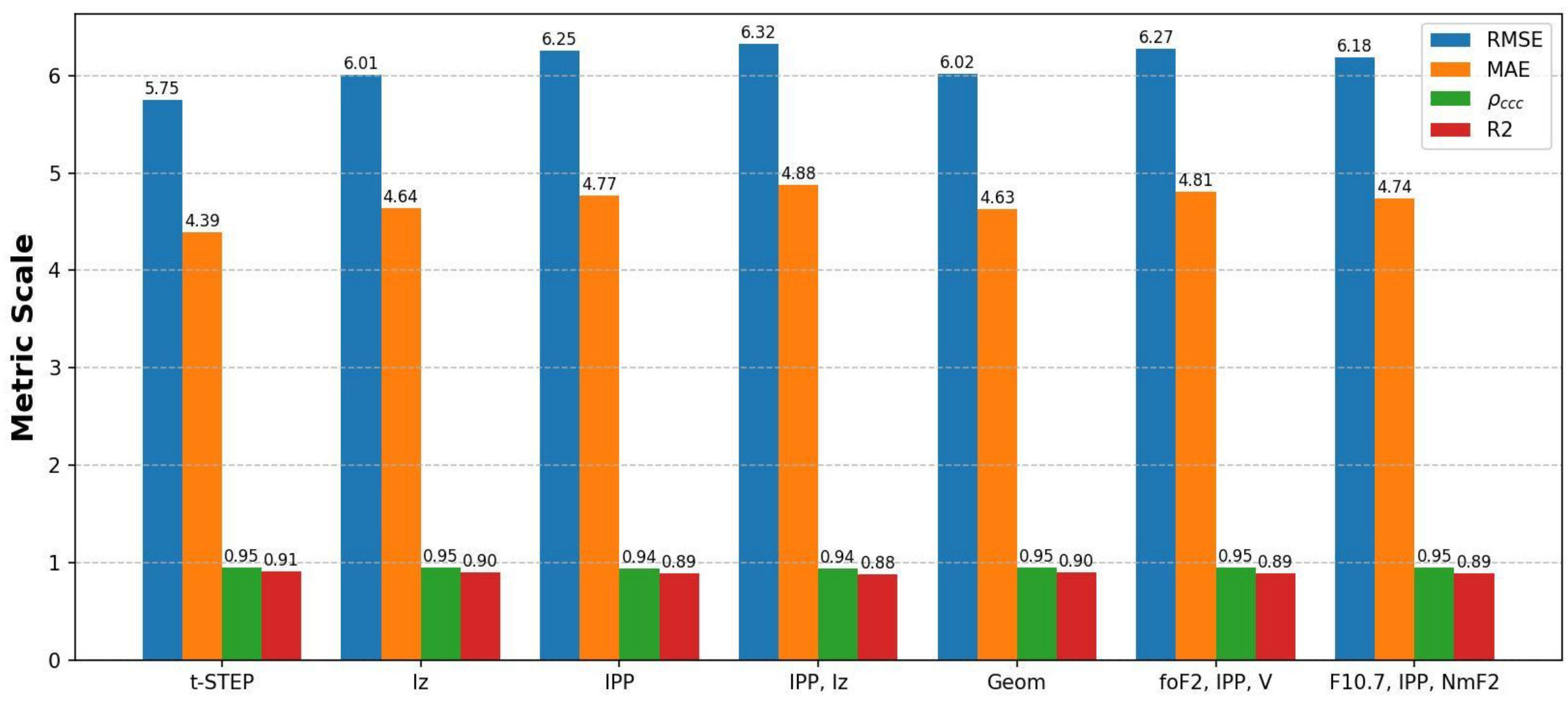

**Fig. 11** Metric scores for six (6) variants of t-STEP constructed by dynamically marginalizing the contributions of individual and feature groups. The plots have been put in the context of the original t-STEP model (leftmost plot) to aid observational evidence

In Fig. 11, we see that model accuracies drop variably with feature labels and group. These drops, compared to the t-STEP, may not be huge, as overall TEC trends were captured. However, when analyzed thoroughly in the context of our second objective, we realized that the effects became non-trivial. The other models either failed to entirely preserve the physical dynamics that generated the embedded transients. A typical example for Iz is shown in Fig. 12. Clearly, the model lacked confidence in describing the electrodynamics of irregularities, which led them to become trapped in the background noise. Subjecting them to high-pass filtering (HPF) indicated their presence, however, the pronouncements do not compare to t-STEP's. These observations confirm that ionosonde-driven parameters serve as pivotal energy constraints for the general physics of irregularity formation. Hence, for any model seeking to forecast irregularities occurrences, incorporating ionosonde observations is essential and highly recommended.

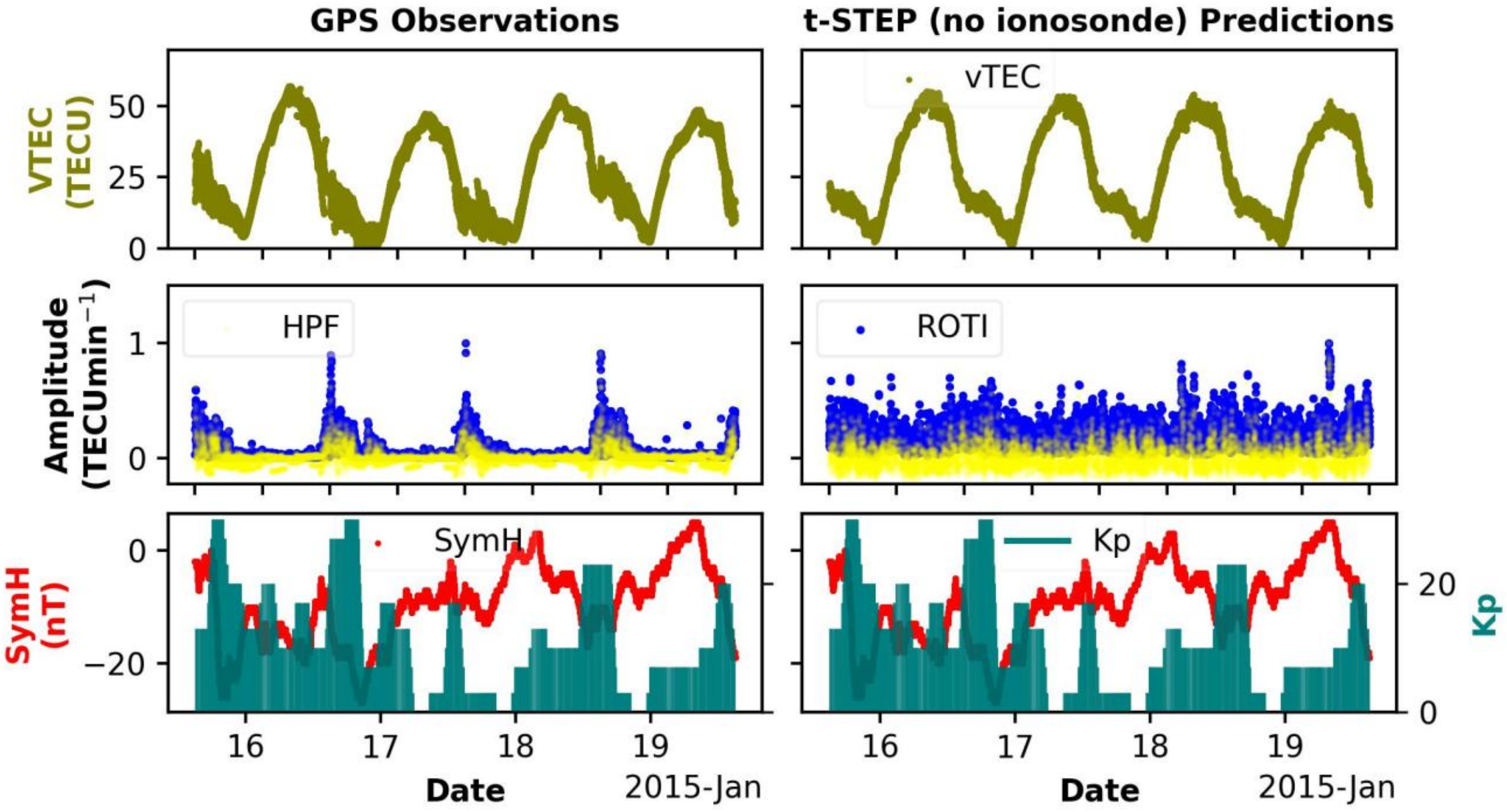

**Fig. 12** A typical model output generated after removing the influence of Iz (the second model in Fig. 11). This example uses the same weak-to-negligible storm interval described in Fig. 6

## SHAP explanations

The SHapley Additive exPlanations (SHAP) is a model-agnostic method for interpreting ML models' outcome. It shows the contribution of each input ensemble to the predicted output. With SHAP, global and local feature contributions can be assessed to aid in spatial and temporal explanations. Here, we show the global contribution of features to the quiet and Patrick's day storm-time predictions in Fig. 6 and Fig. 8. We also examine the temporal evolutions of the storm time TEC values using the SHAP force plots at three time snapshots in sequence.

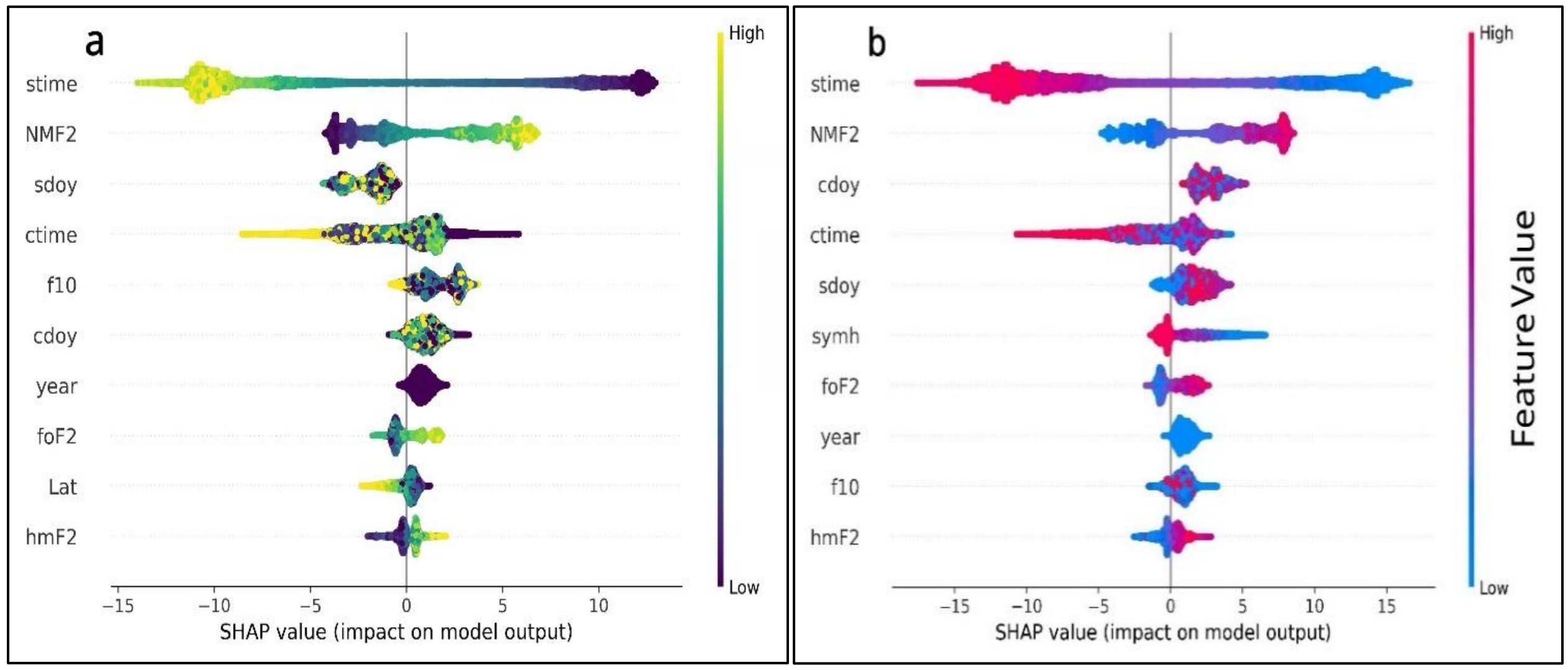

**Fig. 13** SHAP Summary plots of the top 10 important features and their effects on the model under the quiet conditions (a) and disturbed (b) conditions (for Figs. 6 and 8 TEC predictions, respectively)

The summary plot in Fig. 13 shows how features are updated by importance (top to bottom) to influence model predictions. Each dot represents the feature values, while the colors are shapely values corresponding to the feature's state that influenced the high (yellow/red) or low (dark blue/blue) TEC prediction. At first glance, we see that under quiet (a) and disturbed (b) conditions, the sine and cosine of time and Doy (stime, ctime, sdoy, cdoy) and NmF2 are the controlling parameters in the TEC prediction. This is consistent with established theoretical derivations. While time and Doy embeddings controlled diurnal and seasonal variations, NmF2 ensured that electron density changes were captured.

A striking observation which confirmed that t-STEP is rooted in physics and does not make random predictions is the importance ranking levels of the storm time index (SymH) under conditions (a) and (b). During quiet times, the model understood that the disturbance coupling is weak, hence, it relegated SymH below the topmost ranks. The opposite is true for storm conditions where the SymH impact is higher on the importance hierarchy.

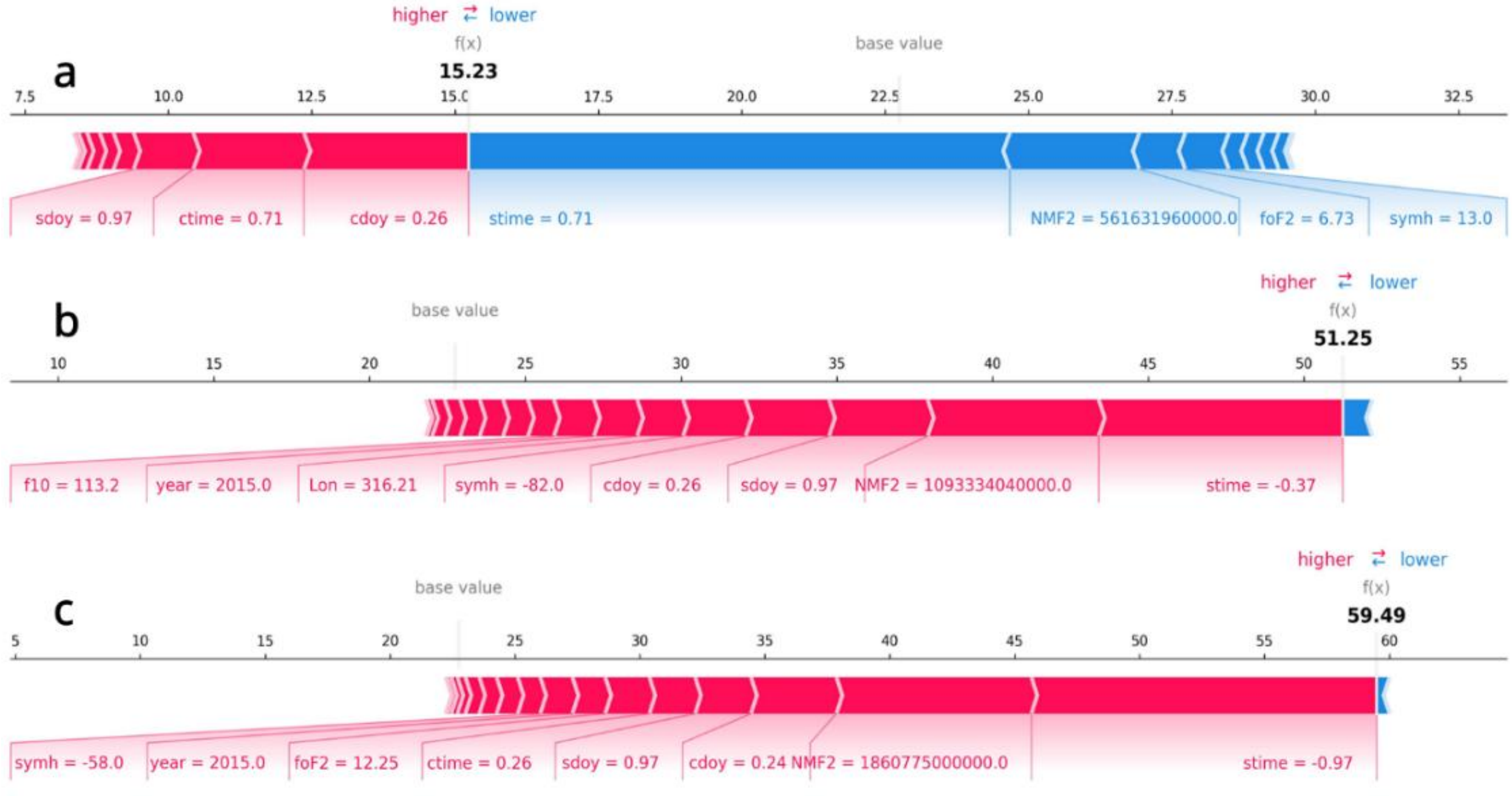

**Fig. 14** SHAP force plot for three time snapshots (a) 2015-03-17 03:00:30, (b) 2015-03-17 13:30:00, and (c) 2015-03-17 19:00:00 during the St. Patrick's day storm

Figure 14 was used to examine t-STEP's decision in predicting individual TEC values. In this graph, the model assumes a global value for TEC, known as "the base value" around which future updates are made as the feature values become fully available to describe the domain space. Then the value moves to the actual prediction f(x). In Fig. 14a, which is a time in the SSC, SymH turns positive (+13 nT). After this, seasonal components tend to influence higher TEC values. Because the Geospace conditions at this time did not fully radiate storm phase conditions, there was a large offset from the dominant diurnal, electron density, and SymH components to mitigate any high TEC prediction. This consequently resulted in the model's output of 15.23 TECU prediction. Figs. 14b and c are excerpts from the excursion and recovery phases of the storm. At this point, the model understood the full state in the Geospace as all features tend to radiate disturbed conditions (red). This consequently increased TEC values even in the recovery phases, as a result of charge depositions into the ionosphere post geomagnetic storm occurrences.

**Comparison of t-STEP with the LSTM model**

In Table 1, we see that the accuracy of t-STEP and LSTM was almost similar for the TEC prediction task. However, as we mentioned earlier, the only way to determine robustness was to subject the model to reconstructing ROTI morphologies within the predicted signal. We have clearly demonstrated with thorough qualitative and quantitative assessments that t-STEP presents ideal predictions and reconstructions to support our objectives. Since benchmarking ROTI reconstructions against the IRI model is impossible, we developed an LSTM deep learning model for these comparisons. The core modeling processes follow the descriptions in the methodology section.

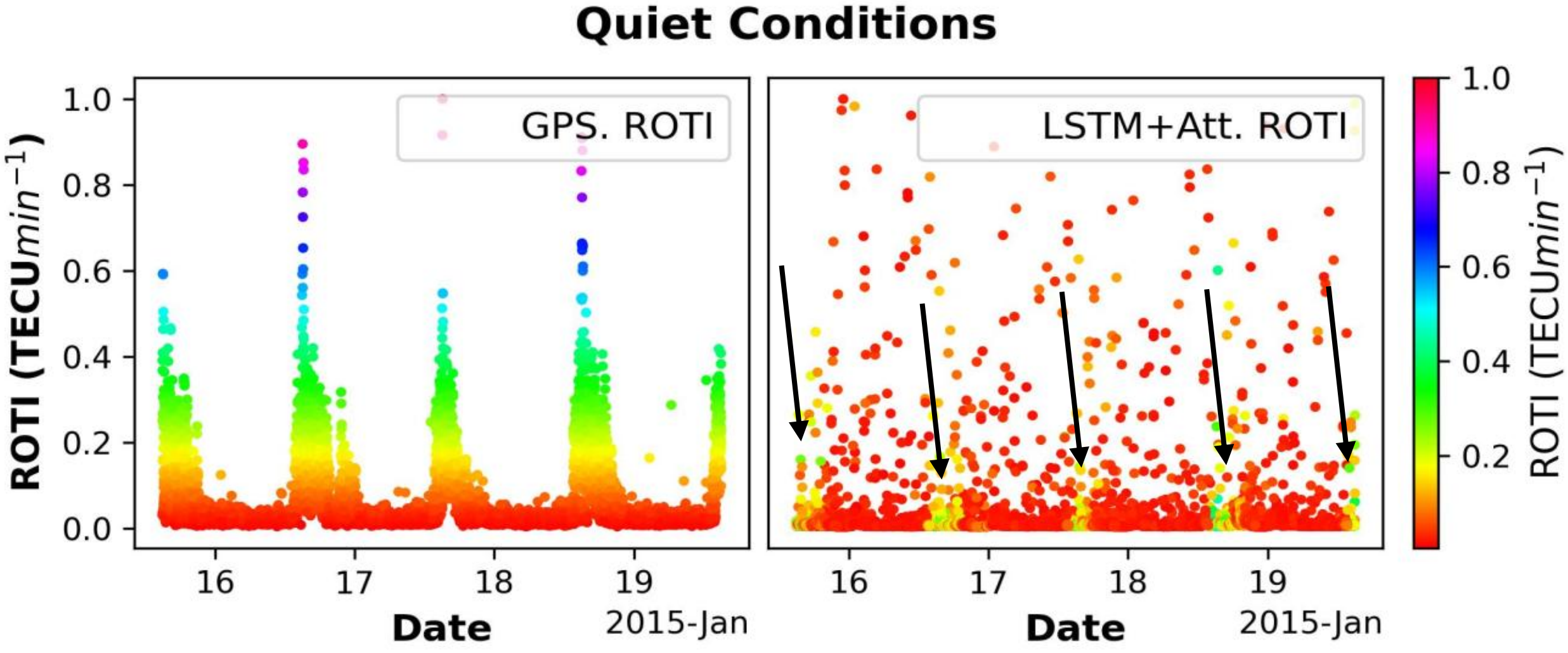

**Fig. 15** ROTI reconstructions from GPS (left) versus LSTM (right) during quiet conditions. The arrows point to the expected temporal positions of ROTI, color-coded with GPS observations

Figure 15 is a typical ROTI reconstructions based on LSTM for the quiet time events discussed earlier in this study. For all instances of the LSTM's irregularity reconstructions, the morphologies were completely distorted. They appeared to be engulfed by noise artifacts. To investigate this further, we analyzed and color-coded the predicted signals irregularities with those from the GPS observations. Even so, no clear patterns were visible. The unclear morphologies suggested that perhaps LSTMs might be too sensitive to fast changing signals but find practical applications in slow changing signals, hence, its remarkable successes in hourly TEC predictions in the literature. We therefore conclude that, for this objective, LSTM, regardless of its attention units, underperforms t-STEP for irregularity retention.

**Comparison of t-STEP and LSTM relative performances**

In Table 2, we summarize the TEC prediction metrics for t-STEP and LSTM under the three space weather scenarios already discussed. The only advantage LSTM demonstrated was its ability to predict TEC in moderate storm conditions better than t-STEP. We identify the competitive advantage with bold fonts in the proceeding tables.

The DTW algorithm is robust at matching and evaluating similar shapes and morphologies within two time varying signals. We applied the algorithm to identify embedded shape matches within the LSTM ROTI signals and evaluated the resulting outcomes using standard performance metrics. The goal was to align similar irregularity patterns that are not clearly discernible from the GPS observations. The performance of perfectly matched patterns is summarized in Table 3 and in Table S2 of the supplementary material.

**Table 2** Comparison of metrics between t-STEP and LSTM for the three space weather conditions

| Model | Metrics | Quiet Condition | Moderate Storm | Strong Storm |
|---|---|---|---|---|
| t-**STEP** | MAE | **3.84** | 8.96 | **5.05** |
| | RMSE | **5.03** | 10.94 | **6.97** |
| | $R^2$ | **0.90** | 0.73 | **0.88** |
| | $\rho_{ccc}$ | 0.95 | 0.84 | 0.94 |
| **LSTM** | MAE | 4.05 | **5.69** | 5.19 |
| | RMSE | 5.27 | **7.31** | 7.12 |
| | $R^2$ | 0.89 | **0.88** | 0.87 |
| | $\rho_{ccc}$ | 0.95 | **0.93** | 0.94 |

**Table 3** Dynamic Time Warping metrics for the ROTI scenarios under t-STEP and LSTM

| **Storm Scenario** | **MAE (TECUmin-1)** | **RMSE (TECUmin-1)** | $\rho_{ccc}$ | $R^2$ |
|---|---|---|---|---|
| **t-STEP** | | | | |
| A (weak to no storm) | 0.019 | **0.033** | **0.87** | **0.72** |
| B (moderate to weak storm) | 0.025 | 0.046 | **0.88** | 0.70 |
| C (strong to moderate storm) | 0.025 | 0.043 | **0.88** | **0.74** |
| **LSTM+ATT** | | | | |
| A (weak to no storm) | **0.016** | 0.040 | 0.84 | 0.64 |
| B (moderate to weak storm) | **0.017** | **0.035** | 0.87 | **0.73** |
| C (strong to moderate storm) | **0.020** | 0.043 | 0.84 | 0.67 |

**Observational perspectives**

Throughout the experiment, we observed the TEC prediction and irregularity reconstruction stability under t-STEP. For a prediction with 30s resolution, we initially anticipated otherwise, especially considering the complexity of the spatio-temporal dynamics of all 32 GPS satellites over the receiver. However, the results proved rewarding, giving us the empirical edge to advance efforts towards small scale events. Based on our observation, we further hypothesize that for chaotic prediction tasks such as this, gradient-boosting ensemble methods appear to outperform their deep learning counterparts. Though consistent with few literature findings including mentions in (Carvalho et al. 2022; Tete et al. 2024), further experiments would be needed to fully validate this hypothesis to help advance prediction tasks of this nature within Earth system sciences.

## Conclusions

In this work, we introduced a system engineered temporal Sampling and TEC Prediction (t-STEP) model that predicts the ionospheric TEC at 30s, a cadence relevant for TEC irregularities estimations. These irregularities have been captured in the rate of TEC index (ROTI) in this study. To aid in the development of t-STEP, we have used TEC observations from a GPS receiver station in the Brazilian sector and across solar cycle 24 (2009 to 2014). Here, we have used t-STEP to demonstrate significant TEC predictions across the high solar activity year of 2015 and assessed these predictions with metrics including $R^2$ and concordance correlations, where 0.91 and 0.95 overall accuracies were achieved. We have demonstrated and evaluated the health of the model's TEC output for irregularity detections and found good reproducibility of large scale gradient irregularities consequential of three main space weather conditions. We evaluated the hourly version of t-STEP in the context of the IRI-2020 outputs. In this experiment, IRI could produce the hourly TEC at about 69% while t-STEP does the same at 93% accuracy, corresponding to about 35% increase in performance and 54% prediction skill. In terms of ROTI reconstructions, we observed that the model's reconstructions matched the GPS observations well, revealing adequate occurrence periodicities. Significant ROTI morphologies were preserved in the two TEC signals (GPS and modeled TEC) signifying the possibility of using a single model to detect ionospheric transients. On the contrary, these irregularities could not be captured clearly in the LSTM model implemented on the same datasets and under the same space weather conditions. The TEC climatology in the LSTM was captured but the gradients were overshadowed with noise artefacts. This suggests that LSTMs might not be adequate for chaotic prediction tasks. It should be noted that the current implementation was limited to a single station. As the current observations looks promising, we are exploring plausible ways to constrain the models to predict TEC to scale to aid the tracking of small scale TEC transients before extending to multiple stations. Another future direction would be to compare against transformer architectures and assess their irregularity preservations across nowcast and forecast scenarios.

## Data Availability

The GPS data used to arrive at the conclusion of this work can be assessed from the Brazilian network, Rede Brasileira de Moni-toramento Contínuo dos sistemas GNSS (RBMC) at: "https://geoftp.ibge.gov.br/informacoes_sobre_posicionamento_geodesico/rbmc/" The ionosonde data was obtained from the EMBRACE network "https://www2.inpe.br/climaespacial/portal/the-embrace-program/" and the DIDBase GIRO web portal "https://giro.uml.edu/ionoweb/" via python interfacing. Solar and geomagnetic activity indices have been accessed using the OMNI module of the Pyspedas framework "https://pyspedas.readthedocs.io/en/latest/". Access to geomagnetic storm information ("https://www.spaceweatherlive.com/en/auroral-activity/top-50-geomagnetic-storms") was done from Space weather live.

## Statements and Declarations

Competing Interests: The authors declare that they have no competing interests.

Funding**:** There was no funding supporting this work.